\documentclass{article}
\usepackage{ijcai24}

\usepackage{times}
\usepackage{soul}
\usepackage{url}
\usepackage[hidelinks]{hyperref}
\usepackage[utf8]{inputenc}
\usepackage[small]{caption}
\usepackage{graphicx}
\usepackage{amsmath}
\usepackage{amsthm}
\usepackage{booktabs}
\usepackage{algorithm}
\usepackage{algorithmic}
\usepackage[switch]{lineno}
\usepackage{color, colortbl}

\definecolor{Lightgray}{gray}{0.85}
\definecolor{Darkgray}{gray}{0.2}
\definecolor{Lightblue}{rgb}{0, 1, 1}
\definecolor{aliceblue}{rgb}{0.94, 0.97, 1.0}
\definecolor{coltable}{rgb}{0.94, 1.0, 1.0} 
\linenumbers
\usepackage{url}            % simple URL typesetting
\usepackage{booktabs}       % professional-quality tables
\usepackage{amsfonts}       % blackboard math symbols
\usepackage{amsmath}
\usepackage{amssymb}
\usepackage{nicefrac}       % compact symbols for 1/2, etc.
\usepackage{microtype}      % microtypography
\usepackage[dvipsnames]{xcolor}         % colors
\usepackage{csquotes}
\usepackage{pifont}

\usepackage{paralist, amsmath, amssymb, graphicx}
\usepackage{amssymb}
\usepackage{mathtools}
\usepackage{nccmath}

\usepackage{nicefrac}
\usepackage{stmaryrd}
\usepackage{bm}
\usepackage{cancel}
\usepackage{multirow}

\usepackage{xfrac}
\usepackage{mathrsfs}%fancy fonts like mscr
\usepackage{amscd}%simple commutative diagrams
\usepackage{dsfont}
\usepackage{tikz}
\usepackage{epstopdf}
\usepackage{enumerate}
\usepackage{bm}
\usepackage{chngcntr} 
\usepackage{makecell}
\usepackage{placeins}
\usepackage{algorithm}
\usepackage{algorithmic}
\usepackage{multicol}
\usepackage{xargs}
\usepackage{appendix}
\usepackage[textwidth=1.5cm,textsize=tiny]{todonotes}
\newcommandx{\improvement}[2][1=]{\todo[linecolor=black,backgroundcolor=cyan,bordercolor=black,#1]{#2}}
\newcommandx{\highlights}[2][1=]{\todo[linecolor=black,backgroundcolor=yellow,bordercolor=black,#1]{#2}}
\usepackage{soul}

\newcommand{\cmark}{\color{Emerald}\ding{51}}
\newcommand{\xmark}{\color{BrickRed}\ding{55}}
\newcommand{\attention}{\color{BurntOrange}\ding{115}}

\newcommand{\green}{\color{Emerald}}

\newcommand{\blue}{\color{SeaGreen}}

\usepackage{comment}

\title{Functional Graph Convolutional Networks: A unified multi-task and multi-modal learning framework to facilitate health and social-care insights}
\author{
Tobia Boschi$^1$
\and
Francesca Bonin$^1$
\and
Rodrigo Ordonez-Hurtado$^1$
\and \\
Cécile Rousseau$^2$ 
\and
Alessandra Pascale$^1$, 
\And 
John Dinsmore$^3$
\affiliations
$^1$IBM Research Europe, Dublin \\
$^2$École Centrale de Lille \\
$^3$Trinity Centre for Practice and Healthcare Innovation, Trinity College of Dublin
\emails
tobia.boschi@ibm.com,
FBonin@ie.ibm.com,
Rodrigo.Ordonez.Hurtado@ibm.com, \\
cecile.rousseau@centrale.centralelille.fr,
apascale@ie.ibm.com,
dinsmorj@tcd.ie
}

\begin{document}
\nolinenumbers
\maketitle

%-=-=-=-=-=-=-=-=-=-=-=-=-=-=-=-=-=-=-=-=-=-=-=-=
%  Abstract
%-=-=-=-=-=-=-=-=-=-=-=-=-=-=-=-=-=-=-=-=-=-=-=-=

\begin{abstract}

This paper introduces a novel Functional Graph Convolutional Network (funGCN) framework that combines Functional Data Analysis and Graph Convolutional Networks
to address the complexities of multi-task and multi-modal learning in digital health and longitudinal studies.
With the growing importance of health solutions to improve health care and social support, ensure healthy lives, and promote well-being at all ages, funGCN offers a unified approach to handle multivariate longitudinal data for multiple entities and ensures interpretability even with small sample sizes. Key innovations include task-specific embedding components that manage different data types, the ability to perform classification, regression, and forecasting, and the creation of a knowledge graph for insightful data interpretation. The efficacy of funGCN is validated through simulation experiments and a real-data application. 
% \vspace{0.3cm}

% This paper introduces a novel Functional Graph Convolutional Network (funGCN) framework that combines Functional Data Analysis and Graph Convolutional Networks to facilitate the analysis of complex digital health and longitudinal studies aiming to ensure healthy lives and promote well-being at all ages.
% funGCN is a multi-modal learning framework that performs different analytical tasks and identifies the most significant interconnections between variables of different natures. It has the potential to provide new insights into personalized health and social care models. Key funGCN innovations include task-specific embedding components that manage different data types, the ability to perform classification, regression, and forecasting, and the creation of a knowledge graph for insightful data interpretation. The efficacy of funGCN is validated through simulation experiments and real-data application experiments. 

\end{abstract}

%-=-=-=-=-=-=-=-=-=-=-=-=-=-=-=-=-=-=-=-=-=-=-=-=
%
%  Introduction
%
%-=-=-=-=-=-=-=-=-=-=-=-=-=-=-=-=-=-=-=-=-=-=-=-=
\section{Introduction}
\label{sec:introduction}

% - UN interest in elderly wellbeing
% - Focus on quality of life prediction -

% The World Health Organization [1] defined quality of life as an “individual perception of his or her living situation, understood in a cultural context, value system and in relation to the objectives, expectations and standards of a given society” (p. 2). From this perspective, health-related quality of life includes areas such as physical health, psychological state, level of independence of the person, personal relationships, beliefs in a particular context or the natural environment, social support, and perceived social support [2–6].

% - Why it is difficult to calculate

% - Need to run trials, trials are costly are constly, but ML offer the possibility to learn from previous data.
% - However, data are multimodal, longitudinal etc... etc.. so classic ML does not exploit all the information.
% - Solution: our prediction.

Ensure healthy lives and promote well-being for all at all ages and leave no one behind – this idea is embedded in the core vision of the UN’s 2030 Agenda for Sustainable Development\footnote{SDG, goal 3, \url{https://sdgs.un.org/goals/goal3}} \cite{resolution2015transforming}.
In line with this, in the contemporary healthcare landscape, digital health solutions and longitudinal studies have become increasingly significant to monitor health care and wellbeing of populations around the world. These approaches enable a comprehensive understanding of health outcomes over time, providing valuable insights into disease progression, treatment efficacy, and overall patients' well-being \cite{garssen2021does}\cite{cuff2023evolution}. Digital health solution are therefore critical tools for achieving the UN sustainable development goals. 

With the advent of new technologies, data are collected from multiple sources, including surveys, phone interviews, wearable devices, and mobile health applications, to cite a few. These tools allow patients to capture a wide array of health-related signals from the comfort of their homes. They facilitate the entry of multiple measurements over time \cite{vijayan2021review} and enhance medical professionals' and policymakers' understanding of a population’s health, the impact of interventions, or, more in general, the well-being of subjects. 
However, they also add to the complexity of the analysis.

Artificial intelligence (AI) plays a key role in effectively dealing with such diverse information. Investigating these studies necessitates a variety of analytical tasks, including \emph{classification}, \emph{longitudinal regression}, and \emph{forecasting}. Classification is crucial for separating patient subgroups or ``arms'', assessing treatments’ effectiveness, or identifying the presence of diseases. Longitudinal regression can estimate critical variables that evolve over the entire study's duration, like patient well-being or healthcare service utilization, and can help determine the impact of a trial or a solution on individual patients. Forecasting, on the other hand, predicts future trends,  informing on a trial's potential success and interventions’ forthcoming effects \cite{hu2015online}.
On top of the different required analytical tasks, the analysis of the collected data presents substantial challenges. First, the variables are of different modalities. Second, the longitudinal signals can be recorded at various times and frequencies. Third, the difficulties of recruiting participants may result in small sample sizes, even smaller than the number of variables \cite{ildstad2001small}. Last, the interpretability of the results is crucial: understanding the relationships between variables and how they influence specific outcomes can enhance future trial designs, improve patient care, and optimize resource allocation \cite{hakkoum2022interpretability}\cite{farah2023assessment}.
All the above challenges constitute a complex methodological barrier, even for the most advanced AI and machine learning (ML) approaches. Hence, there is a need for new informative methodologies able to i) deal with multi-modal data and multivariate longitudinal signals for multiple entities, ii) perform different inference tasks
, and iii) be effective even when the sample size is small. 
In response to these demands, we introduce a novel approach called Functional Graph Convolutional Network (\texttt{funGCN}), which synergizes Functional Data Analysis (FDA) \cite{ramsay2005}\cite{kokoszka2017introduction} and Graph Convolutional Networks (GCNs) \cite{zhang2019graph}. FDA a dynamic area of statistical research that allows working with multivariate longitudinal data by estimating smooth curves across a continuous domain. GCNs are renowned for their capacity to discern relations between variables. By integrating FDA and GCNs, we create a unified framework for multi-task learning with multi-modal longitudinal data. 
The innovative aspects of \texttt{funGCN} include:
% \vspace{-0.4cm}
\begin{list}{$\square$}{\leftmargin=1em \itemindent=0em}
    \item[\footnotesize $\bullet$] The introduction of two task-specific embeddings designed for multi-modality. Specifically, they can handle longitudinal, categorical, and constant scalar variables
    (the latter are numerical values that remain constant over time) and facilitate the comparison of diverse variables and statistical entities throughout the temporal domain.

    % \vspace{-0.2cm}
    \item[\footnotesize $\bullet$] The capability to execute three analytics tasks: classification, (longitudinal) regression, and forecast. Moreover, given the embeddings’ flexibility, \texttt{funGCN} can concurrently perform regression and classification of multiple target variables within a single training session.

    % \vspace{-0.2cm}
    \item[\footnotesize $\bullet$] The generation of a knowledge graph that quantifies the connections between all variables, facilitating dimensionality reduction and providing interpretable insights into the outcomes, even in scenarios with small sample sizes.
\end{list}
The effectiveness of \texttt{funGCN} is demonstrated through simulation experiments and a real-data application, showcasing its ability to construct an informative knowledge graph and offering a new approach for analyzing complex longitudinal datasets.

This work is part of the EU Horizon 2020 SEURO Project\footnote{\href{https://seuro2020.eu/}{https://seuro2020.eu/}} that aims to develop and evaluate a digital social health platform to improve and advance home-based self-management and integrated care for older adults (over 65 years) with multiple chronic conditions.
The project focuses on deploying new technologies for improving health care in elderly population and on evaluating innovative, flexible, and individual-centered healthcare models through data from diverse longitudinal studies.
The project brings together academic/research institutions, Small to Medium Enterprises (SMEs), health service providers and NGOs and benefits from the  a multidisciplinary collaboration between computer scientists/statistician, health-care professionals, stakeholders in the health/social care field.

%-=-=-=-=-=-=-=-=-=-=-=-=-=-=-=-=-=-=-=-=-=-=-=-=
%  Related work
%-=-=-=-=-=-=-=-=-=-=-=-=-=-=-=-=-=-=-=-=-=-=-=-=

\paragraph{Related work.}
Multi-modal GCN, Temporal GCN, and time-series analysis and forecasting are all active research fields. However, aside from \texttt{funGCN}, there seems to be a lack of a comprehensive framework that combines all their capabilities. 
For a clearer understanding of the terminology associated with the applicability of the various models, see Table \ref{tab:definitions}.

Recently work in \cite{langbridge2023causal} introduced
a \emph{Causal Temporal GCN} for multivariate time series, while novel approaches in \cite{jiang2022multi}, \cite{vijay2023tsmixer}, and \cite{zhou2024one} extended \emph{transformers} and \emph{large language models} to time series. Yet, these approaches do not consider multi-modality or multiple statistical entities.
\cite{d2022fusing} presents a multi-modal GCN for classification but, differently from \texttt{funGCN}, does not tackle longitudinal variables. 
\emph{Long short-term memory (LSTM) networks} \cite{hochreiter1997long} and \emph{gated recurrent unit (GRU) networks} \cite{chung2014empirical} allow to perform multiple-task for longitudinal variable and multiple-entities, however they cannot  handle multi-modal data. 
Similarly, classical ML algorithms such \emph{supporting vector machines (SVM)} \cite{muller1997predicting} and \emph{random forest} \cite{breiman2001random}, can manage
multi-modal data only after flattening the longitudinal features, losing the time dependency information and interpretability while, at the same time, increasing the total number of variables.  
On the other hand, FDA has shown promising results in dealing with multi-modal variables across numerous longitudinal applications \cite{ullah2013applications}\cite{cremona2019functional}\cite{boschi2021functional} and, recently, has been effectively extended to deep learning methodologies \cite{rao2023nonlinear}.
\texttt{funGCN} is designed to combine the strengths of the FDA in handling multi-modal longitudinal data with the GCN's capabilities at detecting their interrelationships and producing interpretable outcomes. 
\vspace*{0.15cm}

\noindent
In the remainder of this paper, we first present our methodology (Section \ref{sec:methodology}), then we evaluate the performance of our approach against various competing algorithms through simulations (Section \ref{sec:simulations}), and investigate a real-world application using the longitudinal SHARE dataset (Section \ref{sec:share}). Finally, we draw our conclusions in Section \ref{sec:conclusions}.

\begin{table*}[!t]
    \caption{\textbf{Definitions.} Terminology associated with a model applicability.}
\centerline{
\scalebox{1}{
    \begin{tabular}{l|l}
        \textbf{Term} &  \textbf{Definition} \\
        \Xhline{2\arrayrulewidth}
        \rowcolor{coltable}
       \textit{Multivariate time series} & Multiple longitudinal variables for one entity/subject. \\
        \textit{Multi-modal model} & Model dealing with variables of different modalities (e.g. longitudinal, categorical,...).\\
        \rowcolor{coltable}
        \textit{Multiple entities model}& Model dealing with more statistical entities (e.g. several subjects).\\
        \textit{Multi-task} & Model performing more that one analytical tasks (i.e. regression, classification, forecast).\\
        \Xhline{1.5\arrayrulewidth}
    \end{tabular}
}}
\label{tab:definitions}
\end{table*} 

%-=-=-=-=-=-=-=-=-=-=-=-=-=-=-=-=-=-=-=-=-=-=-=-=
%
%  Methodology
%
%-=-=-=-=-=-=-=-=-=-=-=-=-=-=-=-=-=-=-=-=-=-=-=-=
\section{Methodology}
\label{sec:methodology}

\begin{figure*}[!t]
    \centering
    % \hspace*{-1.5cm}
    \includegraphics[width=0.98\textwidth]{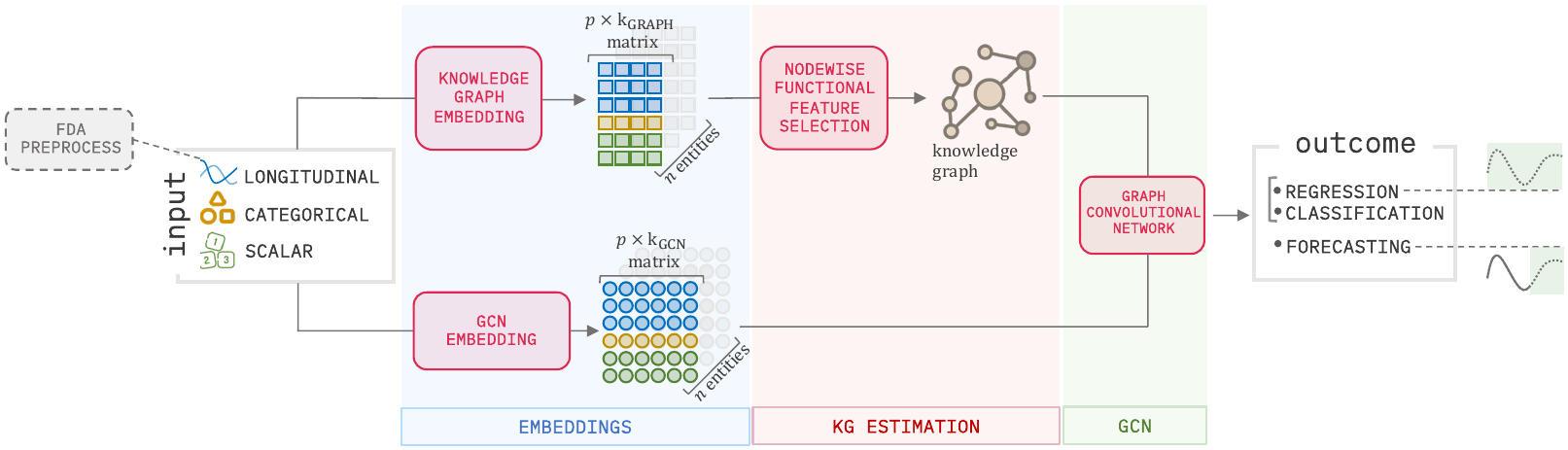}
    % \vspace*{-0.1cm}
    \caption{\textbf{funGCN architecture.} Embedding modules are discussed in Subsection \ref{subsec:embeddings}, the knowledge graph estimation in Subsection \ref{subsec:kg_estimation}, and the GCN module in Subsection \ref{subsec:gcn}.}
    \label{fig:architecture}
\end{figure*}

The architecture of \texttt{funGCN} is illustrated in Figure \ref{fig:architecture}. 
The input data are transformed through two distinct \emph{embedding} modules, each serving a preparatory role — \emph{KG embedding}, estimating the knowledge graph through a node-wise feature selection approach, and \emph{GCN embedding}, preparing the input for the GCN component.
The embedding modules' outputs are tensors of dimensions $(n, p, k_{\text{graph}})$ and $(n, p, k_{\text{gcn}})$, respectively. 
The GCN takes as input both the knowledge graph and its specifically embedded data, and performs three analytics tasks: regression, classification, and forecasting. While regression estimates the feature across the entire time domain, forecasting predicts the \emph{horizon} (future) trajectory of a longitudinal feature given its \emph{history} (past) values. 

The problem dimensions are defined by $n$, representing the number of statistical entities, and $p = p_l + p_c + p_s$, which denotes the total number of variables (or features), where $p_l, p_c, p_s$ indicate the numerosity of each input data modality. Longitudinal variables ($p_l$) evolve over time, categorical variables ($p_c$) can have an arbitrary number of levels, and scalar features ($p_s$) are numerical values that remain constant.

%The input data are transformed through 2 distinct \emph{embedding} modules, each serving a preparatory role — \textbf{Embeddings for KG}: for estimating the knowledge graph through a node-wise feature selection approach and the other for the GCN component.
%The embedding modules output are tensors of dimensions $(n, p, k_{\text{graph}})$ and $(n, p, k_{\text{gcn}})$, respectively. 
%The GCN takes as input the knowledge graph and its specifically embedded data and performs three tasks: regression, classification, and forecasting. While regression estimates the feature across the entire time domain, forecasting predicts the \emph{horizon} (future) trajectory of a longitudinal feature given its \emph{history} (past) values. 

\paragraph{Assumptions on longitudinal features.} For each of the $n$ statistical entities, we observe $p_f$ distinct longitudinal variables, denoted as $\mathcal{X}_{ij}$ for $i=1, \dots, n$ and $j=1, \dots, p_f$.  Typically, in real-world scenarios, these variables are sampled at discrete time points, which might differ across variables and entities \cite{ramsay2005}\cite{ullah2013applications}. 
However, FDA provides various established preprocessing tools to construct smooth and continuous curves from discrete measurements, such as \emph{basis expansions}, \emph{moving average smoothing}, and \emph{sparse conditional estimation} \cite{yao2005functional}.  
Throughout this paper, we assume that such preprocessing has been applied where necessary and  consider $\mathcal{X}_{ij} \in \mathbb{L}^2(\mathcal T)$, the space of square-integrable functions over the closed and bounded interval $\mathcal T$. This implies that if $f$ is a function in $\mathbb L^2(\mathcal{T})$, then
$\lVert f\rVert_{\mathbb{L}^2}^2= \langle f, f\rangle_{\mathbb{L}^2} = \int_\mathcal{T} f^2(t) \text{dt} < \infty$.
Hence, $\mathcal{X}_{ij}$ are infinite-dimensional mathematical objects.

%-=-=-=-=-=-=-=-=-=-=-=-=-=-=-=-=-=-=-=-=-=-=-=-=
%  Embeddings   
%-=-=-=-=-=-=-=-=-=-=-=-=-=-=-=-=-=-=-=-=-=-=-=-=
\subsection{Embedding modules}
\label{subsec:embeddings}

The two embedding modules differ on the representation of longitudinal features, while the same transformation is applied to categorical and scalar variables. 

%-=-=-=-=-=-=-=-=-=-=-=-=-=-=-=-=-=-=-=-=-=-=-=-=
%  Longitudinal features   
%-=-=-=-=-=-=-=-=-=-=-=-=-=-=-=-=-=-=-=-=-=-=-=-=
\paragraph{Longitudinal features.} To transform the longitudinal variable into finite-dimensional objects, we employ a tensor representation, expressing each variable as linear combinations of basis functions \cite{ramsay2005}.
Let $(b^j_1, \dots, b^j_k)$ denote the $k$ functional basis for the feature $j$. Then, the embedded longitudinal features are a tensor $X^l \in \mathbb{R}^{n \times p_l \times k}$ defined as:
\begin{equation*}
    X^l_{ijs} = \langle \mathcal X_{ji}, b^j_s \rangle_{\mathbb{L}^2} = \int_\mathcal{T}  X_{ji}(t) b^j_s(t) \text{dt}.
\end{equation*}
Hence, a functional feature for a specific entity is converted to a vector in $\mathbb{R}^k$, with the coefficients derived by computing the $\mathbb{L}^2$-inner product between the curve and each basis function. The selection of the basis system is crucial, as it influences the characteristics of the resulting embedding. To effectively exploit their properties, we propose two different sets of basis functions for the knowledge graph and the GCN embedding modules.
\vspace*{0.1cm}

\noindent
\emph{KG embedding.}
The longitudinal features are represented using their first $k_{\text{graph}}$ \emph{functional principal components (FPCs)}, derived through an eigen-decomposition of the functional covariance operator \cite{kokoszka2017introduction}.
FPCs capture the directions of larger variability of the curves and are a well-established technique in FDA, generally used to solve a wide range of problems, including regression \cite{reiss2007functional} and classification \cite{wang2016functional}. Given their proven effectiveness in detecting significant relationships among variables \cite{boschi2023fasten}, FPCs are an optimal basis system to determine the structure of the node-wise feature selection module input. 
\vspace*{0.1cm}

\noindent
\emph{GCN embedding.} The longitudinal features are embedded using $k_{\text{gcn}}$ \emph{equispaced cubic B-splines}. B-splines, known for their piece-wise polynomial structure, offer a flexible yet stable method for curves representation \cite{eilers1996flexible}.
Differently from FPCs, which are derived from the data and vary across features, B-spline bases are pre-determined and consistent across different variables. Moreover, while FPCs can influence the entire time domain, cubic B-splines are non-zero over a limited range, defined by a three-knot interval. The localized influence is crucial for effective forecasting while preserving the original embedded form. It enables a clear distinction between the coefficients associated with past values and those associated with the future, which ensures continuity between the history and the horizon prediction and makes B-splines an ideal tool to define the GCN input.

%-=-=-=-=-=-=-=-=-=-=-=-=-=-=-=-=-=-=-=-=-=-=-=-=
%  Categorical features   
%-=-=-=-=-=-=-=-=-=-=-=-=-=-=-=-=-=-=-=-=-=-=-=-=
\paragraph{Categorical features.} Each of the $p_c$ categorical variables, which can have an arbitrary number of levels,  is mapped to a $k$-dimensional space using a standard embedding layer in \texttt{PyTorch} \cite{paszke2019pytorch}.
This process assigns a unique $k$-dimensional vector to every level of the categorical variable, with $k=k_{\text{graph}}, k_{\text{gcn}}$ depending on the embedding module. 
Unlike the typical method where vector representations are learned and optimized within the GCN module, \texttt{funGCN} pre-determines these vectors before the network's initialization.  This approach ensures a static representation that, while not updated during GCN training, allows the integration of multi-modal data by maintaining consistent embeddings.

%-=-=-=-=-=-=-=-=-=-=-=-=-=-=-=-=-=-=-=-=-=-=-=-=
%  Scalar features   
%-=-=-=-=-=-=-=-=-=-=-=-=-=-=-=-=-=-=-=-=-=-=-=-=
\paragraph{Scalar features.} Scalar features are treated as constant functions over time. We follow the same module-specific basis representation used for the longitudinal variables. 
% \vspace*{0.1cm}

\paragraph{}
The embedding modules transform each feature modality into $k_{\text{graph}}$ and $k_{\text{gcn}}$-dimensional vectors, resulting in output tensors $X_{\text{graph}} \in \mathbb{R}^{n \times p \times k_{\text{graph}}}$ and $X_{\text{gcn}} \in \mathbb{R}^{n \times p \times k_{\text{gcn}}}$, respectively. 
These tensors are then standardized to ensure that, if the third dimension is fixed, feature values across all entities have a mean of 0 and a standard deviation of 1. This process guarantees uniform representation of features.
% with dimensions $(n, p, k\graph)$ and $(n, p, k\gcn)$. 
For each entity, the features are encapsulated in a $p \times k$ matrix. This structure aids the comparison of different entities and enables the fusion of different modalities, while preserving the time-dependency information of the longitudinal features.

%-=-=-=-=-=-=-=-=-=-=-=-=-=-=-=-=-=-=-=-=-=-=-=-=
%  Selection of K 
%-=-=-=-=-=-=-=-=-=-=-=-=-=-=-=-=-=-=-=-=-=-=-=-=
\paragraph{Selection of $\mathbf{k_{graph}}$ and  $\mathbf{k_{gcn}}$.}

The choice of $k$ for longitudinal features determines how closely the basis representation approximates the original curves. A lower $k_{\text{graph}}$, such as fewer than 5, offers computational efficiency and effectiveness. Indeed,  FPCs provide a parsimonious yet comprehensive basis system: few components capture most of the curves' variability and identify key feature relationships. On the other hand, B-splines require a larger number of basis functions to reconstruct the original signal. For capturing smooth curves accurately, it is recommended to set $k_{\text{gcn}}$ around 10, with adjustments based on signal complexity and task.

%-=-=-=-=-=-=-=-=-=-=-=-=-=-=-=-=-=-=-=-=-=-=-=-=
%  Knowledge Graph estimation   
%-=-=-=-=-=-=-=-=-=-=-=-=-=-=-=-=-=-=-=-=-=-=-=-=
\subsection{Knowledge Graph estimation}
\label{subsec:kg_estimation}
\nopagebreak  % ROH: I added this to avoid leaving the subsection title separated form the next paragraph.
The aim of the \emph{node-wise functional feature selection module} is the creation of a knowledge graph with nodes representing the $p$ features and edges indicating their association strength. Hence, the graph information can be encapsulated in a $p \times p$ \emph{adjacency matrix} $A$. To estimate $A$, a node-wise regression method is employed, conducting separate feature selection for each feature as a target variable \cite{lee2015learning}. This strategy enhances the accuracy and specificity of the knowledge graph by allowing for individualized analysis of feature relationships.

\begin{algorithm}[!t]
\caption{Node-wise functional feature selection}
\label{alg:nodewise}
\begin{algorithmic}
\vspace{0.1cm}
\STATE \texttt{\textbf{GOAL:}} Estimate adjacency matrix $A \in \mathbb{R}^{p\times p}$ \\
\vspace{0.15cm}
\STATE \texttt{1.} \texttt{SET} $A=0$. \texttt{FOR} $j=1,\dots p$:
\begin{itemize}
    \item Perform \emph{functional feature selection} \\
    \vspace{0.1cm}
    \texttt{SET} target $=j$ and penalty $\lambda = c_{\lambda}\lambda_{\text{max}}$ \\
    \vspace{0.1cm}
    \texttt{SET} $c_{\lambda} = 1$ and active set $\mathcal{S} = \o$
    
    \item[] \texttt{WHILE} $|\mathcal{S}|< p_{\text{max}}$:
    \begin{itemize}
        \item[(a)] \texttt{IF} feature $t$ selected and $t \notin \mathcal{S}$: 
        \begin{itemize}
            \vspace*{-0.1cm}
            \item[$\rightarrow$] add $t$ to $\mathcal{S}$ and set $A_{jt} = c_{\lambda}$ 
        \end{itemize}
        \vspace*{-0.15cm}
        \item[(b)] decrease $c_{\lambda}$
    \end{itemize} 
\end{itemize}
\STATE \texttt{2.} Make $A$ symmetric: $A = \tfrac{1}{2}AA^T$
\vspace{0.15cm}
\STATE \texttt{3.} Prune $A$: if $A_{ij} < \theta \Rightarrow A_{ij} = 0$ and normalize it.

% \vspace*{-0.3cm}
\end{algorithmic}
\end{algorithm}

We adopt the functional feature selection methodology proposed in \cite{boschi2023fasten}. This approach was designed to work with longitudinal features and basis representation. Still, given the flexible structure of $X_{\text{graph}}$, we can enhance its applicability to multi-modal data. 
The node-wise strategy is outlined in Algorithm \ref{alg:nodewise}. For each target variable $j$, $j=1,\dots,p$, we explore different penalty parameters $\lambda = c_\lambda \lambda_{\max}$. Initially, $c_\lambda$ is set to 1, corresponding to 
% $\lambda_{\max}$, indicating 
no active features.  
As $c_\lambda$ gradually reduces, the number of selected features in the model increases. We stop the procedure when this number reaches an upper bound of $p_{\max}$.
In the adjacency matrix $A$, for each row corresponding to a target variable $j$, we mark the columns linked to selected features with their respective $c_\lambda$ values, indicating the order of selection within the model. The diagonal elements are set to 1 for self-association, and all other non-selected feature columns are zeroed out. The earlier a feature is chosen, the higher its $c_\lambda$ value.
%, signifying a stronger association. 
Thus, the association strength reflects the selection sequence of features.

Once the node-wise selection routine is completed, first, we compute the symmetric matrix $\tilde{A}= \tfrac{1}{2}(A+A^T)$, and we then prune the matrix by setting values below a certain threshold $\theta$ to zero. Finally, we normalize it as follows:
\begin{align*}
    \bar A = D\tilde{A}D, \ \ \text{with} \ \  D =  \text{diag}\bigg( \Big( \sum_{j=1}^p  \tilde{A}{ij}\Big)^{-\tfrac{1}{2}}\bigg). 
\end{align*}
This pruning and normalization process further reduces the problem dimension and sharpens the GCN focus on the most significant connections, enhancing the efficiency of training. 

Note that, despite the single feature selection procedure being highly computationally efficient, the graph construction cost grows with the number of features $p$. However, the graph is independent of the specific tasks or targets, and, therefore,  it can be used across different GCN training scenarios without being recomputed. 

%-=-=-=-=-=-=-=-=-=-=-=-=-=-=-=-=-=-=-=-=-=-=-=-=
%  Graph Convolutional Network
%-=-=-=-=-=-=-=-=-=-=-=-=-=-=-=-=-=-=-=-=-=-=-=-=
\subsection{Graph Convolutional Network}
\label{subsec:gcn}

The GCN module has a simple structure consisting of two \emph{convolutional} layers, each followed by a \emph{ReLU} activation function, and a final \emph{linear} layer.

\paragraph{Input.}
The GCN receives as input the preprocessed tensor $X_{\text{gcn}} \in \mathbb{R}^{n \times p \times k_{\text{gcn}}}$ and the adjacency matrix $\bar A$.  At this stage, it is necessary to define one or more target variables, denoted as $p_{\text{target}}$.
In forecasting, $X_{\text{gcn}}$ is divided into 2 parts based on its third dimension. We define $k_2 = r_f*k_{\text{gcn}}$  and $k_1 = k_{\text{gcn}} - k_2$. Here,  $r_f$ represents the ratio of the time domain that will be forecasted, and $k_1$ and $k_2$ denote the number of coefficients associated with history and horizon, respectively. 
We can simply consider $k_1 = k_2 = k_{\text{gcn}}$ for the regression and classification tasks. 

\paragraph{Training.}
We employ the \emph{Adam} optimizer \cite{kingma2014adam} and the \emph{means-square error loss} to deal with finite vectors. 
The $n$ entities are divided into training and validation sets of dimension $n_{\text{train}}$ and $n_{\text{val}}$, respectively.
Throughout each epoch, the training process iterates over the $n_{\text{train}}$ entities feeding the GCN a $p \times k_1$ matrix and receiving a $p_{\text{target}} \times k_2$ output.
The optimization procedure occurs after each epoch.
Monitoring validation loss enables early stopping if no improvement is observed for $v_{\text{stop}}$ consecutive epochs or upon reaching the maximum number of epochs.

\paragraph{Mapping back to original features.}
For longitudinal and scalar variables, their original form can be easily recovered once the training is completed due to the adaptable nature of their embeddings. Given the estimated coefficients $\hat c_{ijs}$ and the basis system $(b^j_1, \dots, b^j_k)$, the curves are computed as follows \cite{kokoszka2017introduction}:
\begin{equation*}
    \hat{\mathcal{X}}_{ij} = \sum_{s=1}^{k_{\text{gcn}}} \hat c_{ijs} b^j_s \,.
\end{equation*}
However, for categorical variables, the embedding cannot be directly inverted to the original space. Instead, a nearest \emph{neighbors algorithm} \cite{cover1967nearest} is employed to match each estimated vector with the nearest original category, ensuring a meaningful approximation of the categorical data.

\paragraph{}
The GCN learns the most important connections between different types of features from the adjacency matrix. Despite its straightforward architecture, it handles multi-modal data and executes different analytics tasks. Notably, $p_{\text{target}}$ can be greater than one, meaning the GCN can predict multiple targets after one training. Furthermore, it is feasible to train the GCN for both classification and regression simultaneously, given that the problem dimensionality is the same for both tasks.

%-=-=-=-=-=-=-=-=-=-=-=-=-=-=-=-=-=-=-=-=-=-=-=-=
%
%  Simulation experiments
%
%-=-=-=-=-=-=-=-=-=-=-=-=-=-=-=-=-=-=-=-=-=-=-=-=
\section{Simulation experiments}
\label{sec:simulations}

\begin{table*}[t!]
\caption{\textbf{Algorithms comparison.} The table displays tasks, capability, and interpretability properties of the proposed \texttt{funGCN} and other (existing) evaluated algorithms. The \emph{task} columns refer to the possibility of performing either regression, classification, or forecast. The \emph{capability} columns refer to the ability to handle i) multi-modal data, ii) multiple entities, or iii) the number of variables being larger than the sample size. 
The \emph{interpretability} columns refer to the ability to estimate i) effect of single feature on the target, ii) importance of the feature for the specific task, or iii) the global relationship between all the variables. 
[Notation: {\green $\star$} indicates that \texttt{funGCN} is the only algorithm able to concurrently perform regression and classification of multiple target variables within just one training process; {\attention} denotes that \texttt{svm} can handle multi-modal data only after flattening the longitudinal variables, losing the time information.]}

\centerline{
\scalebox{0.95}{
    \begin{tabular}{rccccccccc}
    \Xhline{2\arrayrulewidth}
    \multicolumn{1}{l|}{}        
        & \multicolumn{3}{c|}{{\textbf{tasks}}}
        & \multicolumn{3}{c|}{{\textbf{capability}}}
        & \multicolumn{3}{c}{{\textbf{feature interpretability}}} \\
        % \vspace{0.1cm} \\ 
        
    % \Xhline{0.5\arrayrulewidth}
    \multicolumn{1}{l|}{}       
        & \emph{regress}          
        & \emph{classify}       
        & \multicolumn{1}{c|}{\emph{forecast}}        
        % & \emph{multi-task}   
        & \emph{multi-modal}    
        & \emph{multi-entities}    
        & \multicolumn{1}{c|}{$p > n$} 
        & \emph{single effect}
        & \emph{importance}
        & \emph{global relation} \\
    \Xhline{1.5\arrayrulewidth}
    % \hline
    \multicolumn{1}{r|}{\texttt{funGCN}} 
        & \cmark$\star$                 
        & \cmark$\star$                  
        & \multicolumn{1}{c|}{\cmark}                  
        % & \cmark                 
        & \cmark
        & \cmark
        & \multicolumn{1}{c|}{\cmark}        
        & \xmark 
        & \cmark
        & \cmark  \\
    \multicolumn{1}{r|}{\texttt{lstm}}  
        & \cmark                 
        & \cmark                  
        & \multicolumn{1}{c|}{\cmark}                  
        % & \xmark                 
        & \xmark  
        & \cmark
        & \multicolumn{1}{c|}{\cmark}        
        & \xmark 
        & \xmark
        & \xmark  \\ 
    \multicolumn{1}{r|}{\texttt{gru}}  
        & \cmark                 
        & \cmark                  
        & \multicolumn{1}{c|}{\cmark}                
        % & \xmark                 
        & \xmark   
        & \cmark
        & \multicolumn{1}{c|}{\cmark}        
        & \xmark 
        & \xmark 
        & \xmark  \\ 
    \multicolumn{1}{r|}{\texttt{fReg}} 
        & \cmark                 
        & \xmark                  
        & \multicolumn{1}{c|}{\cmark}                    
        % & \xmark                 
        & \cmark  
        & \cmark
        & \multicolumn{1}{c|}{\xmark}        
        & \cmark  
        & \cmark
        & \xmark  \\ 
    \multicolumn{1}{r|}{\texttt{svm}}  
        & \xmark                 
        & \cmark                  
        & \multicolumn{1}{c|}{\xmark}               
        % & \xmark                 
        & \attention  
        & \cmark
        & \multicolumn{1}{c|}{\cmark}        
        & \xmark  
        & \xmark    
        & \xmark  \\ 
    \Xhline{1.5\arrayrulewidth}
    % \\[-0.25cm]
    % \Xhline{1.5\arrayrulewidth}
    % \multicolumn{1}{r|}{\texttt{arimax}}  
    %     & \xmark                 
    %     & \xmark                  
    %     & \multicolumn{1}{c|}{\cmark}               
    %     & \xmark                 
    %     & \cmark  
    %     & \xmark
    %     & \multicolumn{1}{c|}{\xmark}        
    %     & \cmark                    
    %     & \cmark  \\
    % \multicolumn{1}{r|}{\texttt{MGIN}}  
    %     & \xmark                 
    %     & \cmark                  
    %     & \multicolumn{1}{c|}{\xmark}               
    %     & \xmark                 
    %     & \attention  
    %     & \cmark
    %     & \multicolumn{1}{c|}{\cmark}        
    %     & \xmark                    
    %     & \cmark  \\
    % \multicolumn{1}{r|}{\texttt{CTGCN}}  
    %     & \xmark                 
    %     & \xmark                  
    %     & \multicolumn{1}{c|}{\cmark}               
    %     & \xmark                 
    %     & \xmark  
    %     & \xmark
    %     & \multicolumn{1}{c|}{\cmark}        
    %     & \xmark                    
    %     & \cmark  \\
    % \multicolumn{1}{r|}{\texttt{FPT}}  
    %     & \xmark                 
    %     & \cmark                  
    %     & \multicolumn{1}{c|}{\xmark}               
    %     & \xmark                 
    %     & \attention  
    %     & \xmark
    %     & \multicolumn{1}{c|}{\attention}        
    %     & \xmark                    
    %     & \xmark  \\
    % \Xhline{2\arrayrulewidth}
    \end{tabular}
}
}
\label{tab:comparison} 
\end{table*}

\begin{figure*}[!t]
    \centering
    % \hspace*{-1.5cm}
    \includegraphics[width=1\textwidth]{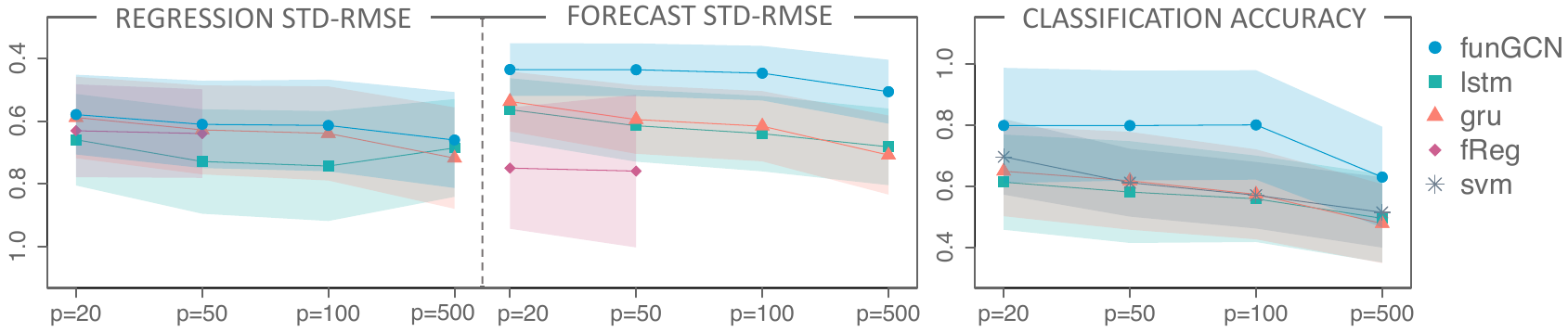}
    % \vspace*{-0.1cm}
    \caption{\textbf{Simulation results.} For each investigated scenario $(p=20, 50, 100, 500)$, dots represent the \emph{average} over 80 replications (20 for each target), and the width of the interval indicates the \emph{standard deviation}. The $y$-$axis$ for \emph{regression} and \emph{forecast} are inverted: higher values correspond to better performance in all the panels.}
    \label{fig:sim_results}
\end{figure*}

\begin{table}[!t]
\caption{\textbf{Simulation CPU time.} Average CPU processing time, in seconds, computed over 80 replications (20 for each target), and across all possible tasks an algorithm performs in each instance. For \texttt{funGCN} we report the time required to create the knowledge graph $({\texttt{kg}})$ and the time needed to complete the inference task $({\texttt{inf}})$. All the computations were executed on a MacBookPro 2021 with an M1 Max processor and 32GB of RAM.
}
\vspace{-0.2cm}
\centerline{
\scalebox{0.88}{
    \begin{tabular}{r|rr|rrrr}
    % \Xhline{2\arrayrulewidth}
    $p$   
        & \texttt{funGCN}$_{\texttt{kg}}$
        & \texttt{funGCN}$_{\texttt{inf}}$
        & \texttt{lstm} 
        & \texttt{gru} 
        & \texttt{fReg} 
        & \texttt{svm} \\ 
    \Xhline{2\arrayrulewidth}
    \texttt{20}  
        & 0.8  
        & 2.5             
        & 4.3   
        & 10.8   
        & 472.5     
        & 0.1    \\
    \texttt{50}  
        & 1.8
        & 2.2          
        & 9.7    
        & 18.6  
        & 5,006.5     
        & 0.2    \\
    \texttt{100} 
        & 3.5        
        & 2.4          
        & 16.8   
        & 25.1  
        & $-$    
        & 0.4   \\
    \texttt{500} 
        & 16.1           
        & 3.1        
        & 80.7   
        & 80.8    
        & $-$    
        & 4.7  \\
    \Xhline{2\arrayrulewidth}
\end{tabular}
}}
\label{tab:sim_time} 
\end{table}

%-=-=-=-=-=-=-=-=-=-=-=-=-=-=-=-=-=-=-=-=-=-=-=-=
%  Methods comparison
%-=-=-=-=-=-=-=-=-=-=-=-=-=-=-=-=-=-=-=-=-=-=-=-=
\paragraph{Methods comparison.}
This section assesses the \texttt{funGCN} model's effectiveness with synthetic data, comparing it to \texttt{lstm}, \texttt{gru}, \texttt{svm}, and functional regression (\texttt{fReg}). The implementation and hyper-parameters used for all the methods are detailed in Section \ref{sec:appendix_hyperparameters} of the Appendix. 
To the best of our knowledge, these are the only methodologies able to handle multivariate longitudinal data and multiple entities. 

Table \ref{tab:comparison} compares the selected methods based on their task execution capabilities, applicable scenarios, and interpretability.
Notably, \texttt{funGCN} is the only framework supporting diverse tasks for multi-modal data and offering comprehensive insights into feature relationships.
 \texttt{lstm} and \texttt{gru} lack interpretability and support for categorical features, necessitating that we remove such variables before executing these methods. 
\texttt{svm} is limited to classification and requires one-hot encoding of categorical features and flattening of longitudinal variables, resulting in the loss of the time-dependency information and interpretability.
Finally, \texttt{fReg} cannot handle classification and scenarios with more features than entities.

%-=-=-=-=-=-=-=-=-=-=-=-=-=-=-=-=-=-=-=-=-=-=-=-=
%  Experiments settings.
%-=-=-=-=-=-=-=-=-=-=-=-=-=-=-=-=-=-=-=-=-=-=-=-=
\paragraph{Experiments settings.}
We set the number of entities $n=300$ and examine four distinct scenarios with varying numbers of features: $p=20, 50, 100, 500$. In each scenario, the proportion of longitudinal, categorical, and scalar variables is $0.6$, $0.2$, and $0.2$, respectively, and the number of interconnected features is $p_0=10$. 
We split $p_0=10$ into four longitudinal, 4 categorical (2 with two levels, 1 with three levels, and 1 with four levels), and 2 scalar features. 
The synthetic data generation process is described in Section \ref{sec:appendix_data_generation} of the Appendix.

In each scenario, the target variables are the 4 longitudinal and the 4 categorical from the $p_0$ interconnected features. For forecasting, we set the forecast ratio at 0.3, utilizing 70$\%$ of the domain for predictions.. For each target, we conduct 20 replications using 225 entities for training and 75 for testing across all methods, totaling 80 replications per task. For \texttt{fReg}, we limit to 10 replications per target in the scenario with 50 features due to its computational complexity (see Table \ref{tab:sim_time}) and do not consider scenarios with 100 and 500 features due to its inability to deal with more parameters than entities. 

Regression and forecasting tasks are assessed using the \emph{standardized root mean-squared error}, which for the longitudinal target $\mathcal{Y}$ is computed as:
\begin{equation*}
    \text{std-RMSE}= \bigg( \frac{1}{n} \sum_{i=1}^n \frac{1}{\text{sd}(\mathcal{Y}_i)} \int_\mathcal{T} \big(\mathcal{Y}_i - \hat{\mathcal{Y}}_i \big)^2 \text{dt} \bigg)^{\tfrac{1}{2}}.
\end{equation*}
Classification performances are evaluated in terms of \emph{accuracy}, reflecting the proportion of correctly predicted instances.

%-=-=-=-=-=-=-=-=-=-=-=-=-=-=-=-=-=-=-=-=-=-=-=-=
%  Results
%-=-=-=-=-=-=-=-=-=-=-=-=-=-=-=-=-=-=-=-=-=-=-=-=
\paragraph{Results.}
The results of the simulation experiments are presented in Figure \ref{fig:sim_results}. For each task across varying $p$, the dots illustrate the average performance derived from 80 replications (20 for each target), with the width of the intervals indicating the standard deviation. 
In regression tasks, the performance of all algorithms is comparable, though \texttt{funGCN} exhibits a marginal advantage. In forecasting and classification, \texttt{funGCN} outperforms its competitors, which achieve similar results among themselves. 
These findings underline the versatility and effectiveness of \texttt{funGCN} across different tasks and numbers of features, even in challenging scenarios where the sample size is small relative to the number of variables.

Table \ref{tab:sim_time} shows the average CPU time required for completing all replications across various tasks performed by each algorithm. The computation time for \texttt{funGCN} is divided into two parts: the creation of the knowledge graph and the training of the GCN module. Although the graph was recomputed for each simulation, it is important to note that this computation is required only once in real-world applications, as the graph does not need to be recomputed at inference time. 
\texttt{svm} is the most time-efficient algorithm, whereas \texttt{fReg} is significantly more computationally expensive. Compared to other deep learning methods, \texttt{funGCN} exhibits greater efficiency, particularly as the number of variables increases. Notably, for scenarios with $p=500$, \texttt{funGCN}'s inference phase even surpasses \texttt{svm} in speed, highlighting its efficiency in handling a large number of variables.

%-=-=-=-=-=-=-=-=-=-=-=-=-=-=-=-=-=-=-=-=-=-=-=-=
%
%  SHARE analysis 
%
%-=-=-=-=-=-=-=-=-=-=-=-=-=-=-=-=-=-=-=-=-=-=-=-=
\section{SHARE analysis}
\label{sec:share}

\begin{figure}[!t]
    \centering
    % \hspace*{-1.5cm}
    \includegraphics[width=0.48\textwidth]{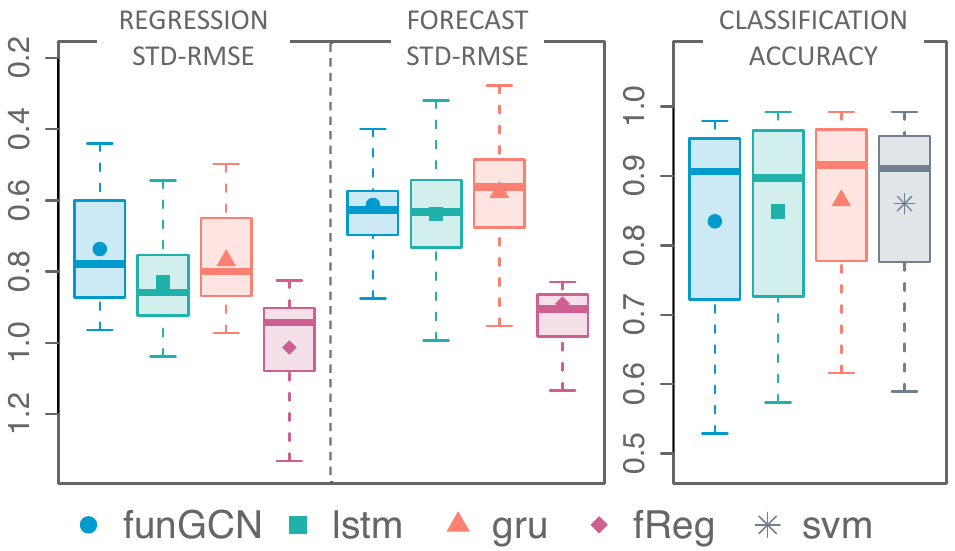}
    % \vspace*{-0.4cm}
    \caption{\textbf{SHARE results.} Boxplots generated from the distribution obtained from a total of 260 replications for regression and forecasting tasks, and 80 replications for classification -- encompassing 20 replications for each target.
    The dots and the lines indicate the means and medians of the distributions, respectively. The $y$-$axis$ for \emph{regression} and \emph{forecast} are inverted: higher values correspond to better performance in all the panels.}
    \label{fig:share_results}
\end{figure}

\begin{figure*}[!t]
    \centering
    % \hspace*{-1.5cm}
    \includegraphics[width=1\textwidth]{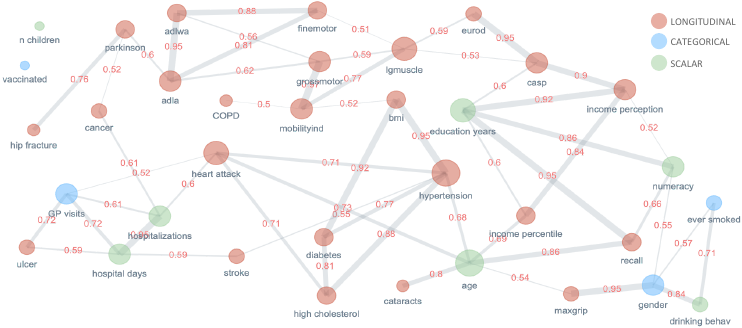}
    % \vspace*{-0.1cm}
    \caption{\textbf{SHARE knowledge graph.} Graph constructed considering all the $1,518$ subjects with $k_{graph} = 3$ and pruning parameter $\theta =0.5$. Each node represents a feature, while different colors refer to different modalities. The nodes' diameter and edges' width are proportional to the connections’ number and strength, respectively.}
    \label{fig:kg_share}
\end{figure*}

The Survey of Health, Ageing and Retirement in Europe (SHARE) is a pivotal research project that examines the impact of health, social, economic, and environmental policies on the life course of European citizens \cite{borsch2013data}\cite{bergmann2017survey}\cite{borsch2020survey}.
SHARE stands out for its longitudinal design, tracking the same individuals across eight survey waves from 2004 to 2020. 
We restrict our analysis to the 1,518 participants who were part of at least seven waves, ensuring sufficient measurements for reliable curve estimation.
We investigate a subset of 35 variables from the EasySHARE dataset \cite{gruber2014generating}, a preprocessed version of the SHARE data.
These variables, which can be longitudinal, categorical, or scalar, span physical and mental health, socio-demographics, and healthcare, offering a rich dataset for comprehensive analysis — the complete list of the examined variables is reported in Appendix Table \ref{tab:share_var_list}. 

Following the strategy in \cite{boschi2024new}, we employ cubic B-splines for smoothing longitudinal variables, with knots at survey dates and a roughness penalty on the curve's second derivative. The optimal smoothing parameter is determined by minimizing the generalized cross-validation criterion \cite{craven1978smoothing}. Despite variations in survey dates and data points across participants, this approach offers a consistent functional representation, allowing for a natural imputation of missing values and facilitating comparisons across the entire timeline.

%-=-=-=-=-=-=-=-=-=-=-=-=-=-=-=-=-=-=-=-=-=-=-=-=
%  Results
%-=-=-=-=-=-=-=-=-=-=-=-=-=-=-=-=-=-=-=-=-=-=-=-=
\paragraph{Across task results.}
To evaluate the algorithms’ performance across the 3 different tasks, we select each of the 13 longitudinal and 4 categorical variables from the chosen 35-feature subset as target variables (detailed in Appendix Table \ref{tab:share_var_list}). We conduct 20 replications for each target, randomly splitting the 1,518 subjects into a training set of 1,139 and a test set of 379.
The aggregate average results are depicted in Figure \ref{fig:share_results}, while the individual targets performance are available in Appendix Table \ref{tab:individual_share_results}. 
%The findings provide limited insights. 
Except for \texttt{fReg}, which shows lower efficacy, all methods — including \texttt{funGCN} — yield similar results across all evaluation metrics and tasks in this analysis.
However, \texttt{funGCN} provides interpretable results, analyzed in the next section.

%-=-=-=-=-=-=-=-=-=-=-=-=-=-=-=-=-=-=-=-=-=-=-=-=
%  Graph Analysis 
%-=-=-=-=-=-=-=-=-=-=-=-=-=-=-=-=-=-=-=-=-=-=-=-=
\paragraph{Graph Analysis.}
While the evaluation metrics are comparable across algorithms, \texttt{funGCN} provides a unique interpretability feature. Figure \ref{fig:kg_share} illustrates the knowledge graph generated for all 1,518 subjects with $k_{graph} = 3$ and pruning parameter $\theta = 0.5$. The graph unveils significant insights into the interconnection between multi-modal features, and also highlights complex relationships between health, social, and economic factors.

In particular, the graph finds connections among chronic conditions such as hypertension, heart attacks, and diabetes, demonstrating their associations with factors like body mass index (BMI) and age. This visualization highlights the diseases’ intertwined nature and potential common drivers or consequences, findings confirmed by medical literature \cite{petrie2018diabetes}\cite{powell2021obesity}.

Moreover, \texttt{funGCN} reveals how variables describing healthcare service utilization, such as the number of general practitioner visits and hospitalizations, are interconnected. This helps to provide a clearer understanding of the healthcare engagement and utilization patterns among individuals with different health conditions \cite{walker2014relationship}.

Noteworthy is also the relationship between \emph{casp}, a quality of life measure, and \emph{eurod}, an index assessing depression levels, alongside perceptions of income and years of education. This connection underscores the multifaceted impacts of mental health, socioeconomic status, and education on overall life satisfaction \cite{sivertsen2015depression}\cite{zanin2017education}.

Additionally, the graph spotlights the link between Parkinson's disease, hip fractures, and variables related to the subjects' mobility, such as \emph{adla} (which measures the activities of daily living) and fine motor skills. This association points to the broader implications of such health conditions on individuals' independence and daily functioning \cite{tan2012relationships}.

Through these detailed insights, the knowledge graph serves as a tool for understanding specific health and social connections and emphasizes the value of \texttt{funGCN} in offering a comprehensive overview of the factors influencing health outcomes.

%-=-=-=-=-=-=-=-=-=-=-=-=-=-=-=-=-=-=-=-=-=-=-=-=
%
%  Conclusions
%
%-=-=-=-=-=-=-=-=-=-=-=-=-=-=-=-=-=-=-=-=-=-=-=-=
\section{Conclusions}
\label{sec:conclusions}

The proposed Functional Graph Convolutional Network (\texttt{funGCN}) model offers a powerful and innovative approach to analyzing complex multi-modal longitudinal data. By combining Functional Data Analysis and Graph Convolutional Networks, \texttt{funGCN} provides a unified framework for multi-task learning and generates interpretable insights through the creation of a knowledge graph. The promising simulation experiments and a real-data application demonstrate \texttt{funGCN}'s capability to improve the analysis of outcomes from the implementation of digital health solutions and longitudinal studies.

\texttt{funGCN} is an essential component of the EU Horizon 2020 SEURO Project  that aims to enhance home-based care and self-management for older and frail adults. 
By identifying the most significant interconnections between variables of different natures, \texttt{funGCN} has the potential to provide new insights on personalized health and social care models. This can assist medical professionals and policymakers in enhancing patient well-being, reducing healthcare costs, and associated burden on patients services.

Future plans include expanding \texttt{funGCN}'s applicability to a wider range of healthcare datasets and settings, incorporating more data types like imaging and genomics, and increasing model interpretability with explainable AI techniques.

%-=-=-=-=-=-=-=-=-=-=-=-=-=-=-=-=-=-=-=-=-=-=-=-=
%
%  Acknowledgments 
%
%-=-=-=-=-=-=-=-=-=-=-=-=-=-=-=-=-=-=-=-=-=-=-=-=
\section*{Acknowledgments}

% To be included in the camera-ready upon acceptance. 
The authors received funding from the European Union’s Horizon 2020 research and innovation program under grant agreement No. 945449.
% This document reflects the views only of the authors, and the European Union cannot be held responsible for any use which may be made of the information contained therein.
The authors also thank their partners in IMEC for helping access the SHARE data.

%-=-=-=-=-=-=-=-=-=-=-=-=-=-=-=-=-=-=-=-=-=-=-=-=
%
%  Bibliography
\bibliographystyle{named}
\bibliography{bib}

%-=-=-=-=-=-=-=-=-=-=-=-=-=-=-=-=-=-=-=-=-=-=-=-=
%
%  Appendix  
%
%-=-=-=-=-=-=-=-=-=-=-=-=-=-=-=-=-=-=-=-=-=-=-=-=
\newpage
\appendix
\onecolumn

\counterwithin{figure}{section}
\counterwithin{table}{section}
\renewcommand\thefigure{\thesection\arabic{figure}}
\renewcommand\thetable{\thesection\arabic{table}}

\begin{center}
    \LARGE \textbf{Appendix}
\end{center}

%-=-=-=-=-=-=-=-=-=-=-=-=-=-=-=-=-=-=-=-=-=-=-=-=
%
%  Algorithms's implementation and hyper-parameters
%
%-=-=-=-=-=-=-=-=-=-=-=-=-=-=-=-=-=-=-=-=-=-=-=-=
\section{Algorithms' implementation and hyper-parameters}
\label{sec:appendix_hyperparameters}

\texttt{funGCN} is developed in \texttt{python}. 
We set $k_{\text{graph}} = 3$, and $k_{\text{gcn}} = 5, 10, 20$ for classification, regression, and forecasting, respectively. The larger $k_{\text{gcn}}$ for forecasting accommodates the division into historical and future-oriented coefficients.
For node-wise regression, we utilize \texttt{fasten}, a functional feature selection approach detailed in Boschi et al. (2023), accessible at:
{\blue \texttt{\href{https://github.com/tobiaboschi/FASTEN}{https://github.com/tobiaboschi/FASTEN}}}. 
The selection process halts once 5 variables form the active set for each target variable. The pruning threshold $\theta$ for the adjacency matrix is set at 0.7 for synthetic data and 0.5 for SHARE data.
The GCN module, implemented in \texttt{PyTorch}, consists of two convolutional layer, each with 32 hidden units. The learning rate for the Adam optimizer is adjusted to $5e$-$5$ for regression and $1e$-$4$ for forecasting and classification. The 20$\%$ of the training set is used as validation set. The early stopping criterion interrupts the training process after 5 epochs without validation loss improvement, capping at a maximum of 50 epochs. \\

\noindent
For the \texttt{lstm}, \texttt{gru}, \texttt{fReg}, and \texttt{svm} models, longitudinal features are evaluated on a grid of 100 uniformly spaced time points for the simulations scenarios, and 192 uniformly spaced time points (one for each month) for the SHARE application. \texttt{lstm}, \texttt{gru} do not take categorical features as input. \texttt{lstm} and \texttt{gru} models exclude categorical features. For \texttt{svm}, categorical features undergo one-hot encoding, and longitudinal features are flattened. \\

\noindent 
\texttt{lstm} and \texttt{gru} are implemented in \texttt{PyTorch} and the network architecture is the same as the GCN module, consisting of two layers with 32 hidden units each. 
The same \texttt{funGCN} trainig procedure is implemented. The Adam optimizer learning rate is set at $1e$-$3$ for \texttt{lstm}, and $1e$-$4$ for \texttt{gru}. For the classification task, a \emph{cross-entropy loss} is employed. \\

\noindent 
The \texttt{SVM} model is utilized from the \texttt{scikit-learn Python package} \cite{pedregosa2011scikit}, with its default settings and an RBF (radial basis function) kernel. \\

\noindent 
The \texttt{fReg} model performs the concurrent regression procedure implemented in the \texttt{refund R package} \cite{crainiceanu2013package}. Following the default parameters, we allocate ten coefficients for each longitudinal feature. In forecasting scenarios, the future values of the longitudinal target serve as the response variable, with past values incorporated into the predictors. Note, this approach does not ensure a continuous transition between observed curves and their future forecasts.

\noindent

%-=-=-=-=-=-=-=-=-=-=-=-=-=-=-=-=-=-=-=-=-=-=-=-=
%
%  Syhnthetic data generation
%
%-=-=-=-=-=-=-=-=-=-=-=-=-=-=-=-=-=-=-=-=-=-=-=-=
\section{Syhnthetic data generation}
\label{sec:appendix_data_generation}

Following other work on functional regression 
\cite{parodi2018simultaneous}, we draw each longitudinal feature $\mathcal {X}_{j}$ from a zero-mean Gaussian process with a Matern covariance function \cite{cressie1999classes} of the form
\begin{equation*}
\small
\begin{split}
        C(t,s) = \frac{\eta^2}{\Gamma(\nu)2^{\nu-1}}\Bigg(\frac{\sqrt{2\nu}}{l} \lvert t-s \rvert\Bigg)^\nu K_\nu\Bigg(\frac{\sqrt{2\nu}}{l} \lvert t-s \rvert \Bigg) \ ,
\end{split}
\end{equation*}
where $K_\nu$ is a modified Bessel function. We employ point-wise variance $\eta^2=1$, range $l=0.25$, and smoothness parameter $\nu=3.5$. Each curve is evaluated at 100 time-points.

\noindent
The scalar variables are sampled from a normal distribution with 0 mean and standard deviation equal to 1. 
For creating a categorical feature, we first determine the number of categories randomly, choosing between two and five. Subsequently, we randomly allocate observations to these categories, aiming for an even distribution across each class. \\

\noindent
To create the 10 interconnected features, we start with 6 longitudinal features and add smoothed correlated noise. The noise is derived from a multivariate normal distribution characterized by a mean vector of 0 and a covariance matrix with entries exceeding 0.4, which is randomly generated. We smooth this noise using a Gaussian filter before integrating this noise with the features. Among these correlated longitudinal features, 2 are transformed into scalar variables by simply taking the average of each curve. 
The categorical variables are created as follows: 
\begin{itemize}
    \item[1.] We construct an n-dimensional vector by evaluating a linear combination of the 4 correlated longitudinal features at randomly selected points. The weights for this combination are drawn from a uniform distribution between $-3$ and $3$.
    \item[2.] These vectors are then normalized to scale the values to fall within a range from 1 up to the total number of categories designated for the variable.
    \item[3.] We convert these scaled values into integers through rounding. Each integer is associated to a category.
\end{itemize}

%-=-=-=-=-=-=-=-=-=-=-=-=-=-=-=-=-=-=-=-=-=-=-=-=
%
%  Additional tables
%
%-=-=-=-=-=-=-=-=-=-=-=-=-=-=-=-=-=-=-=-=-=-=-=-=
\clearpage
\section{Additional tables}
\label{sec:appendix_tables}

\begin{table}[!h] 
%  \footnotesize
%  \small 
%  \centering
  \caption{List of the variables analyzed within the SHARE application. 
  The letter adjacent to the variable name denotes whether it is \emph{longitudinal} (l), \emph{scalar} (s), or \emph{categorical} (c). The letter (a) denotes a scalar variable obtained taking an average across the waves where the values were available. The asterisk indicates variables that have been considered as targets in the models described in Section 
  \ref{sec:share}.
  For more detailed information, the reader should consult the SHARE project website: \texttt{\href{https://share-eric.eu/}{https://share-eric.eu/}}}
%   \hspace*{2.1cm}
%   \hspace*{1.7cm}
  \vspace{0.3cm}
  \centerline{
  \scalebox{1}{
  \begin{tabular}{r|l}
    \textbf{variable} & \textbf{short description} \\
    \Xhline{2\arrayrulewidth}
    \texttt{CASP$^*$ (l)} & quality of life index  \\ 
    \texttt{max grip$^*$ (l)} & maximum of grip strength measure  \\
    \texttt{recall test$^*$ (l)} & number of words recalled in the first trial  \\     
	\texttt{bmi$^*$ (l)} & body mass index  \\     
	\texttt{adlwa$^*$ (l)} & activities of daily living index \\
	\texttt{adla$^*$ (l)} & sum of five daily activities \\
	\texttt{lgmuscle$^*$ (l)} & large muscle index \\
	\texttt{mobilityind$^*$ (l)} & mobility index \\
	\texttt{grossmotor$^*$ (l)} & grossmotor skills index \\
	\texttt{finemotor$^*$ (l)} & finemotor skills index \\
	\texttt{income perception$^*$ (l)} & household able to make ends meet \\
	\texttt{eurod$^*$ (l)} & depression index with EURO-D scale \\
	\texttt{income percentile$^*$ (l)} & household income percentiles \\
	\texttt{heart attack (l)} & 1 if the subject ever had the disease, 0 o.w.\\
	\texttt{high cholesterol (l)} & 1 if the subject ever had the disease, 0 o.w. \\
	\texttt{stroke (l)} & 1 if the subject ever had the disease, 0 o.w. \\
	\texttt{diabetes (l)} & 1 if the subject ever had the disease, 0 o.w. \\
	\texttt{COPD (l)} & 1 if the subject ever had the disease, 0 o.w. \\
	\texttt{cancer (l)} & 1 if the subject ever had the disease, 0 o.w. \\
	\texttt{ulcer (l)} & 1 if the subject ever had the disease, 0 o.w. \\
	\texttt{parkinson (l)} & 1 if the subject ever had the disease, 0 o.w. \\
	\texttt{cataracts (l)} & 1 if the subject ever had the disease, 0 o.w. \\
	\texttt{hip fracture (l)} & 1 if the subject ever had the disease, 0 o.w. \\
	\texttt{age (s)} & age of the subject \\
	\texttt{education years (s)} & years of education \\
	\texttt{n children (a)} & number of children that are still alive   \\
	\texttt{numeracy (a)} & mathematical performance    \\
	\texttt{drinking behav (a)} & times a patient drunk in the last 6 months  \\
	\texttt{hospital days (a)} & days spent at the hospital in the last 6 months  \\
	\texttt{hospitalizations (a)} & number of hospitalizations in the last 6 months  \\
        \texttt{GP visits$^*$ (c)} & number of doctor visits within the study: high, medium, low \\
	\texttt{gender$^*$ (c)} & female or male \\
	\texttt{vaccinated$^*$ (c)} &  being vaccinated during childhood\\
	\texttt{ever smoked$^*$ (c)} & ever smoked daily \\
  \Xhline{2\arrayrulewidth}
  \end{tabular}}}
  \label{tab:share_var_list} 
\end{table}

\begin{table}[!t]
\caption{\textbf{SHARE results}. Average of the task-specific evaluation criterion across 20 replications for each target variable}
% \vspace{-0.2cm}
\centerline{
\scalebox{1}{
    \begin{tabular}{rrrrr}
    \multicolumn{1}{r}{\texttt{\textbf{regression std rmse}}} & \texttt{funGCN}  & \texttt{lstm} & \texttt{gru} & \texttt{fReg}  \\ 
    \Xhline{2\arrayrulewidth}
    \multicolumn{1}{r|}{\texttt{CASP}}                & 0.78 & 0.83 & 0.77 & 0.90 \\ 
    \multicolumn{1}{r|}{\texttt{max grip}}            & 0.55 & 0.89 & 0.83 & 0.86 \\
    \multicolumn{1}{r|}{\texttt{recall}}              & 0.87 & 0.92 & 0.88 & 0.86 \\
    \multicolumn{1}{r|}{\texttt{bmi}}                 & 0.91 & 0.96 & 0.90 & 0.93 \\
    \multicolumn{1}{r|}{\texttt{adlwa}}               & 0.51 & 0.63 & 0.57 & 1.24 \\
    \multicolumn{1}{r|}{\texttt{adla}}                & 0.52 & 0.69 & 0.60 & 1.31 \\
    \multicolumn{1}{r|}{\texttt{lgmuscle}}            & 0.79 & 0.85 & 0.78 & 0.95 \\
    \multicolumn{1}{r|}{\texttt{mobilityind}}         & 0.63 & 0.64 & 0.59 & 1.01 \\
    \multicolumn{1}{r|}{\texttt{grossmotor}}          & 0.63 & 0.78 & 0.66 & 1.04 \\
    \multicolumn{1}{r|}{\texttt{finemotor}}           & 0.72 & 0.79 & 0.75 & 1.24 \\
    \multicolumn{1}{r|}{\texttt{income perception}}   & 0.83 & 0.87 & 0.81 & 0.90 \\
    \multicolumn{1}{r|}{\texttt{eurod}}               & 0.88 & 0.92 & 0.87 & 0.92 \\
    \multicolumn{1}{r|}{\texttt{income percentile}}   & 0.91 & 0.96 & 0.91 & 0.94 \\
    \Xhline{1.5\arrayrulewidth}
    \\[0.2cm]
    
    \multicolumn{1}{r}{\texttt{\textbf{forecast std rmse}}} & \texttt{funGCN}  & \texttt{lstm} & \texttt{gru} & \texttt{fReg}  \\ 
    \Xhline{2\arrayrulewidth}
    \multicolumn{1}{r|}{\texttt{CASP}}                & 0.63 & 0.54 & 0.47 & 0.88  \\ 
    \multicolumn{1}{r|}{\texttt{max grip}}            & 0.28 & 0.37 & 0.30 & 0.59  \\
    \multicolumn{1}{r|}{\texttt{recall}}              & 0.61 & 0.62 & 0.56 & 0.87 \\
    \multicolumn{1}{r|}{\texttt{bmi}}                 & 0.45 & 0.43 & 0.38 & 0.58  \\
    \multicolumn{1}{r|}{\texttt{adlwa}}               & 0.73 & 0.83 & 0.75 & 1.05 \\
    \multicolumn{1}{r|}{\texttt{adla}}                & 0.73 & 0.89 & 0.84 & 1.03 \\
    \multicolumn{1}{r|}{\texttt{lgmuscle}}            & 0.62 & 0.65 & 0.58 & 0.89 \\
    \multicolumn{1}{r|}{\texttt{mobilityind}}         & 0.61 & 0.63 & 0.56 & 0.87 \\
    \multicolumn{1}{r|}{\texttt{grossmotor}}          & 0.69 & 0.71 & 0.65 & 0.92 \\
    \multicolumn{1}{r|}{\texttt{finemotor}}           & 0.75 & 0.81 & 0.76 & 1.01 \\
    \multicolumn{1}{r|}{\texttt{income perception}}   & 0.56 & 0.55 & 0.50 & 0.89 \\
    \multicolumn{1}{r|}{\texttt{eurod}}               & 0.67 & 0.65 & 0.60 & 0.93 \\
    \multicolumn{1}{r|}{\texttt{income percentile}}   & 0.58 & 0.55 & 0.49 & 1.00 \\
    \Xhline{1.5\arrayrulewidth}
    \\[0.2cm]
    
    \multicolumn{1}{r}{\texttt{\textbf{classification accuracy}}}  & \texttt{funGCN}  & \texttt{lstm} & \texttt{gru} & \texttt{svm}  \\ 
    \Xhline{2\arrayrulewidth}
    \multicolumn{1}{r|}{\texttt{GP visits}}          & 0.87 & 0.89 & 0.90 & 0.90 \\
    \multicolumn{1}{r|}{\texttt{gender}}             & 0.93 & 0.89 & 0.94 & 0.92 \\
    \multicolumn{1}{r|}{\texttt{vaccinated}}         & 0.97 & 0.98 & 0.98 & 0.98 \\
    \multicolumn{1}{r|}{\texttt{ever smoked}}        & 0.58 & 0.62 & 0.64 & 0.63 \\
    \Xhline{1.5\arrayrulewidth}
    \end{tabular}
}
}
\label{tab:individual_share_results} 
\end{table}

\end{document}